\def\year{2022}\relax
\documentclass[letterpaper]{article} % DO NOT CHANGE THIS
\usepackage{aaai22}  % DO NOT CHANGE THIS
\usepackage{times}  % DO NOT CHANGE THIS
\usepackage{helvet}  % DO NOT CHANGE THIS
\usepackage{courier}  % DO NOT CHANGE THIS
\usepackage[hyphens]{url}  % DO NOT CHANGE THIS
\usepackage{graphicx} % DO NOT CHANGE THIS
\usepackage{natbib}  % DO NOT CHANGE THIS AND DO NOT ADD ANY OPTIONS TO IT
\usepackage{caption} % DO NOT CHANGE THIS AND DO NOT ADD ANY OPTIONS TO IT
\DeclareCaptionStyle{ruled}{labelfont=normalfont,labelsep=colon,strut=off} % DO NOT CHANGE THIS
\usepackage{algorithm}
\usepackage{algorithmic}
\usepackage[table,xcdraw]{xcolor}
\usepackage{subfigure}
\usepackage{newfloat}
\usepackage{listings}
\floatstyle{ruled}
\newfloat{listing}{tb}{lst}{}
\floatname{listing}{Listing}

\title{Cross-Modal Object Tracking: Modality-Aware Representations and\\ A Unified Benchmark}
 \author {
     Chenglong Li\textsuperscript{\rm 1},
     Tianhao Zhu\textsuperscript{\rm 2}, 
     Lei Liu\textsuperscript{\rm 2}, 
     Xiaonan Si\textsuperscript{\rm 2}, 
     Zilin Fan\textsuperscript{\rm 2}, 
     Sulan Zhai\textsuperscript{\rm 3}
 }
 \affiliations {
     \textsuperscript{\rm 1} School of Artificial Intelligence, Anhui University, Hefei 230601, China\\
     \textsuperscript{\rm 2} School of Computer Science and Technology, Anhui University, Hefei 230601, China
\\
     \textsuperscript{\rm 3} School of Mathematical Sciences, Anhui University, Hefei 230601, China\\
     lcl1314@foxmail.com, zhutianhao79@gmail.com, liulei970507@163.com, 01044@ahu.edu.cn%\{1065682519,1198791575\}@qq.com, 01044@ahu.edu.cn
 }
\begin{document}
% \def\AAAISubNumber{7321}
% \title{Cross-Modal Object Tracking: Modality-Aware Representations and\\ A Unified Benchmark}
% \author{Paper ID \AAAISubNumber}
\maketitle

\begin{abstract}
In many visual systems, visual tracking often bases on RGB image sequences, in which some targets are invalid in low-light conditions, and tracking performance is thus affected significantly. Introducing other modalities such as depth and infrared data is an effective way to handle imaging limitations of individual sources, but multi-modal imaging platforms usually require elaborate designs and cannot be applied in many real-world applications at present. Near-infrared (NIR) imaging becomes an essential part of many surveillance cameras, whose imaging is switchable between RGB and NIR based on the light intensity. These two modalities are heterogeneous with very different visual properties and thus bring big challenges for visual tracking. However, existing works have not studied this challenging problem. In this work, we address the cross-modal object tracking problem and contribute a new video dataset, including $654$ cross-modal image sequences with over $481$K frames in total, and the average video length is more than $735$ frames. To promote the research and development of cross-modal object tracking, 
we propose a new algorithm, which learns the modality-aware target representation to mitigate the appearance gap between RGB and NIR modalities in the tracking process.
It is plug-and-play and could thus be flexibly embedded into different tracking frameworks.
Extensive experiments on the dataset are conducted, and we demonstrate the effectiveness of the proposed algorithm in two representative tracking frameworks against $17$ state-of-the-art tracking methods. We will release the dataset for free academic usage, dataset download link and code will be released soon.
%and have submitted the code of the proposed trackers in the supplementary materials.
\end{abstract}

\section{Introduction}
Visual tracking is an important problem in the field of computer vision and plays a critical role in many visual systems, such as visual surveillance,  intelligent transportation, and robotics. However, existing tracking methods often base on RGB image sequences which are sensitive to illumination variations, and some targets are thus invalid in low-light conditions. In such scenarios, the tracking performance of existing methods might degrade significantly.

Some works introduce other modalities such as depth and infrared data to overcome imaging limitations of RGB source~\cite{Song13iccv,li2016learning,li2019rgb}. However, multi-modal imaging platforms usually require elaborate design and cannot be applied in many real-world applications at present. For example, the depth sensors can provide valuable additional depth information to improve tracking results by robust occlusion and model drift handling, but suffer from the limited range (e.g., $4$-$5$ meters at most) and indoor environment~\cite{Song13iccv,li2016learning}. Thermal sensors are usually independent of RGB ones and their visual properties are very different. Therefore, a lot of efforts are needed in platform design and frames alignment~\cite{li2016learning,li2019rgb}.

\begin{figure}[t]
	\centering
    \includegraphics[width=.95\linewidth]{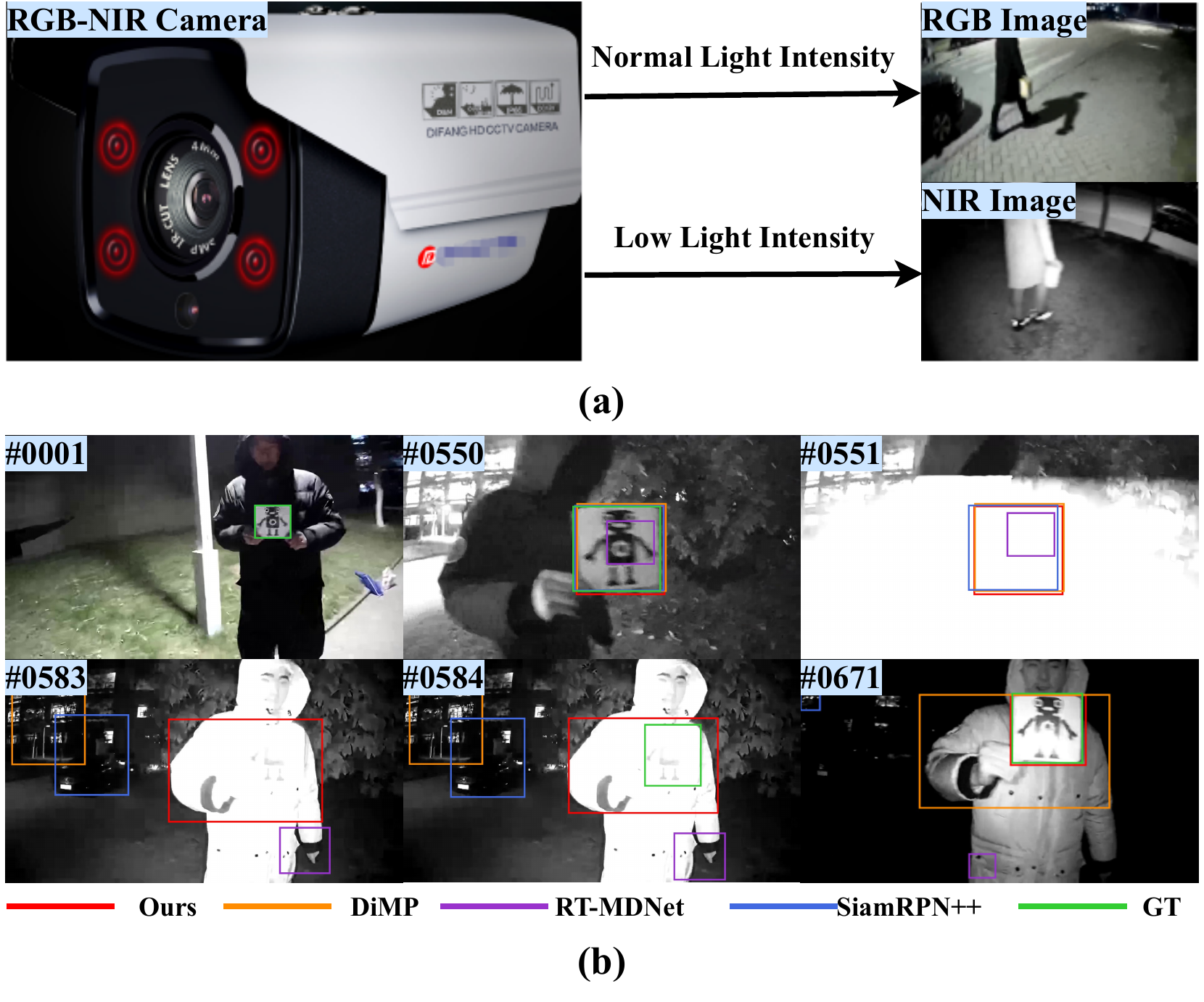}
	%\end{center}
	\caption{(a) Illustration of heterogeneous properties between RGB and NIR modalities. The visual camera changes RGB imaging to NIR when the light intensity becomes low from normal, and vice versa. (b) A comparison of our approach with state-of-the-art trackers in facing the challenge of modality switch, including DiMP~\cite{bhat2019learning}, SiamRPN++~\cite{li2019siamrpn++} and RT-MDNet~\cite{jung2018real}. The results show that our method handles this challenge well but the others trackers fail when the appearance of the target varies significantly caused by modality switch.}
	\label{fig::motivation}
\end{figure}

Near-infrared (NIR) imaging becomes an essential part of many surveillance cameras, whose imaging is switchable between RGB and NIR based on the light intensity, as shown in Fig.~\ref{fig::motivation}(a). This kind of imaging system well handles imaging limitations of RGB source in low-light conditions while avoiding the imaging and platform problems introduced by existing multi-modal visual systems. From Fig.~\ref{fig::motivation}(b) we can also observe that these two modalities are heterogeneous with very different visual properties and the appearance of the target object is thus totally different in different modalities. Such an appearance gap brings big challenges for visual tracking, and existing tracking works have not studied this challenging problem.

In this work, we address the problem of cross-modal object tracking and aim to answer the following two questions. How to design a suitable algorithm, which could mitigate the appearance gap between RGB and NIR modalities and flexibly embedded into different tracking frameworks, for robust cross-modal object tracking? How to create a video benchmark dataset for the promotion of research and development of cross-modal object tracking? %How much do the switch times affect tracking performance?

% \begin{figure*}[!htp]
% 	\begin{center} % 图片居中
% 		%  \fbox{\rule{0pt}{2in} \rule{.9\linewidth}{0pt}}
% 		\includegraphics[width=\linewidth,trim=0 0 0 0,clip]{figure/example.eps}
% 	\end{center}
% 	\caption{A typical examples in our CMOTB. For clarity, we present some key frames in the sequence \emph{Red doll-1}. The ground truths are indicated by red bounding boxes.}
% 	\label{fig::example}
% \end{figure*}

First, we propose a {\bf M}odality-{\bf A}ware cross-{\bf M}odal {\bf O}bject {\bf T}racking algorithm (MArMOT),
%we propose a new plug and play module named {\bf C}ross-{\bf M}odal {\bf M}odality-{\bf A}ware {\bf M}odule (MArMOT), 
which learns the modality-aware target representations to mitigate the appearance gap between RGB and NIR modalities in the tracking  process. MArMOT is plug-and-play and could thus be flexibly embedded into different tracking frameworks. MArMOT includes two parallel CNN branches to learn modality-specific target representations using different sets of training samples. Besides, we do not know which modality appears in the tracking process. Thus, we design a ensemble module to adaptively incorporate effective features from both branches with any modality as input. In this way, the appearance gap between RGB and NIR modalities can be well addressed.

Second, to build a unified benchmark dataset, we collect $654$ cross-modal object tracking sequences. The total number of video frames reaches over $481$K, and the average video length and the maximum length of one sequence are more than $735$ and $1.6$K frames. This dataset contains most of the real-world challenges in cross-modal object tracking task. Most importantly, it contains more challenges in adverse environmental conditions, as shown in Fig.~\ref{fig::motivation}(b), which easily triggers modality switch and significantly declines the capability of visual trackers.

%Finally, we perform performance evaluation of tracking algorithms under different subsets, the division of which is based on the switch times of two modalities. We find that the performance of trackers almost drops with the increase of switch times and the drop is significant when the switch times are more than three. It suggests that the modality switch is a critical challenging factor in cross-modal object tracking.

The major contributions of this work can be summarized as follows. First, we introduce {\bf a new task} called cross-modal object tracking that is very challenging but practical in many visual systems. Second, we propose {\bf a novel algorithm} to mitigate the appearance gap of target object between different modalities for robust cross-modal object tracking, and integrate it into {\bf two typical tracking frameworks} for effectiveness and generality validations. Third, we develop {\bf a three-stage learning algorithm} to train the proposed tracking networks efficiently and effectively. Fourth, we create {\bf a unified benchmark dataset} which contains most of the real-world challenges in cross-modal object tracking.  Finally, we carry out {\bf an extensive experiment} to demonstrate the effectiveness of the proposed approaches against the state-of-the-art trackers and clarify the research room on the cross-modal object tracking. 

\section{MArMOT Trackers}
In this section, we first introduce the proposed {\bf M}odality-{\bf A}ware cross-{\bf M}odal {\bf O}bject {\bf T}racking model (MArMOT), and then the tracking architectures with MArMOT including how embed the proposed plug-and-play MarMOT into two typical tracking frameworks. At last, the three-stage learning algorithm and the tracking details are provided.
% then the three-stage learning algorithm and the tracking details are provided.

\begin{figure}[t]  %[]
	\centering
	%\begin{center} % 图片居中
	%  \fbox{\rule{0pt}{2in} \rule{.9\linewidth}{0pt}}
	\includegraphics[width=.7\linewidth]{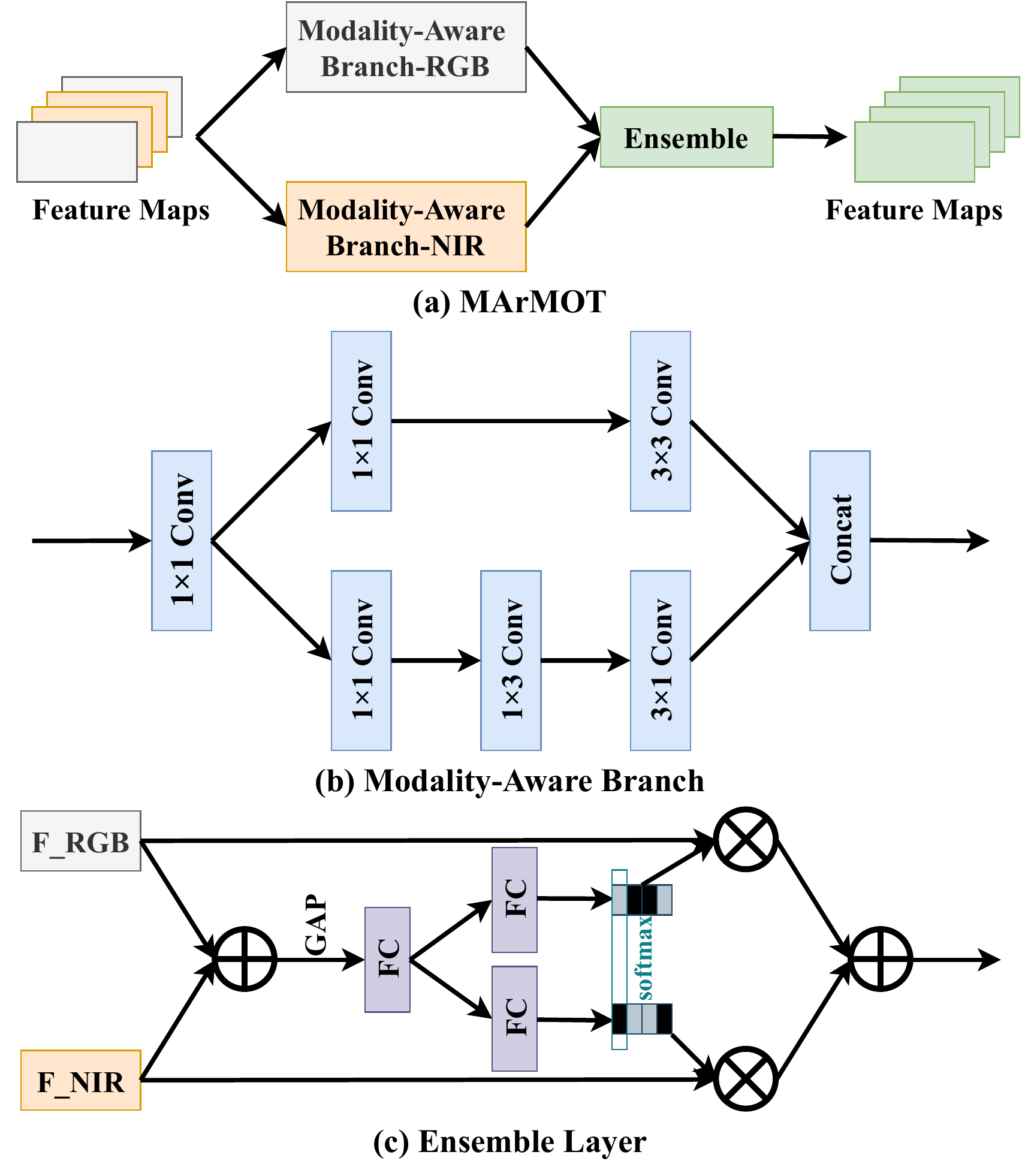}
	%\end{center}
	\caption{Details of MArMOT. The BN+ReLU layers after each Conv layer and the ReLU layers after each FC layer are omitted for clarity. Herein, $\oplus$ and $\otimes$ denote the operations of element-wise addition and multiplication respectively. F\_RGB and F\_NIR denote the output features of RGB and NIR after two parallel modality-aware branches respectively. GAP indicates the global average pooling.}
	\label{fig::sub_network}
\end{figure}

\subsection{MArMOT Model}
In the task of cross-modal object tracking, two modalities are heterogeneous with very different visual properties and thus bring big challenges for visual tracking. To solve this problem, we propose a new MArMOT which learns the modality-aware target representations to mitigate the appearance gap between RGB and NIR modalities in tracking process. Note that MArMOT is plug-and-play and can thus be flexibly embedded into different tracking frameworks.%, as shown in Fig.~\ref{fig::overall_network}.

MArMOT includes two parallel modality-aware branches to learn modality-specific target representations using different sets of training samples. Besides, we do not know which modality appears in the tracking process. Thus, we design an ensemble module to adaptively incorporate effective features from both branches with any modality as input. In this way, the appearance gap between RGB and NIR modalities can be well addressed, as shown in Fig.~\ref{fig::sub_network}.
%Next, we will describe the detailed design of MArMOT.

{\flushleft \bf Modality-Aware Branch}.
%As some trackers~\cite{jung2018real} adopted, we select a lightweight CNN to extract modality-specific features for the tracking task.
%First, we use a VGG network as the base network to extract features for both modalities, and the kernel sizes of three convolutional layers are  $7 \times 7$,  $5 \times 5$, and $3 \times 3$ respectively. 
%After that, we design a two-stream CNN to extract RGB and NIR representations respectively in a parallel way.
%As for the architecture of each branch, we use the inception-like network~\cite{szegedy2016inception} for the effective and efficient computation. The details can be seen in Fig.~\ref{fig::network}.
Two parallel modality-aware branches are followed by backbone network, and used for learning modality-specific representations of the target in different modalities. As for the architecture of each branch, we use the inception-like network~\cite{szegedy2016inception} for the effective and efficient computation. The details can be seen in Fig.~\ref{fig::sub_network}(b). In each branch, the first $1 \times 1$ convolutional layer is used to capture modality-specific representations. Then it is divided into two flows by using another two $1 \times 1$ convolutional layers with half channels to decrease the dimensionality of the input feature and fed into two types of $3 \times 3$ convolution to increase the adaptability of the network to targets of different scales. Their outputs are concatenated together as the modality-specific representation.

{\flushleft \bf Ensemble Layer}.
Due to the particularity of cross-modal object tracking, we design two parallel modality-aware branches to capture the modality-specific representations. However, in tracking process, we only have one modality as the input in each frame and do not know which modality is presented. 
To handle this problem, we design an ensemble layer to adaptively integrate features outputted from two branches given one modal input. By this way, we can obtain the effective features no matter which modality is input.
Specifically, we utilize the SKNet~\cite{li2019selective}  to fuse the features of the two parallel branches by weighting them using the normalized weights, and thus achieve adaptive fusion of these two branch features. The detailed design can be found in Fig.~\ref{fig::sub_network}(c).

\begin{figure}[]  %[!htp]
	\centering
	\includegraphics[width=.7\linewidth]{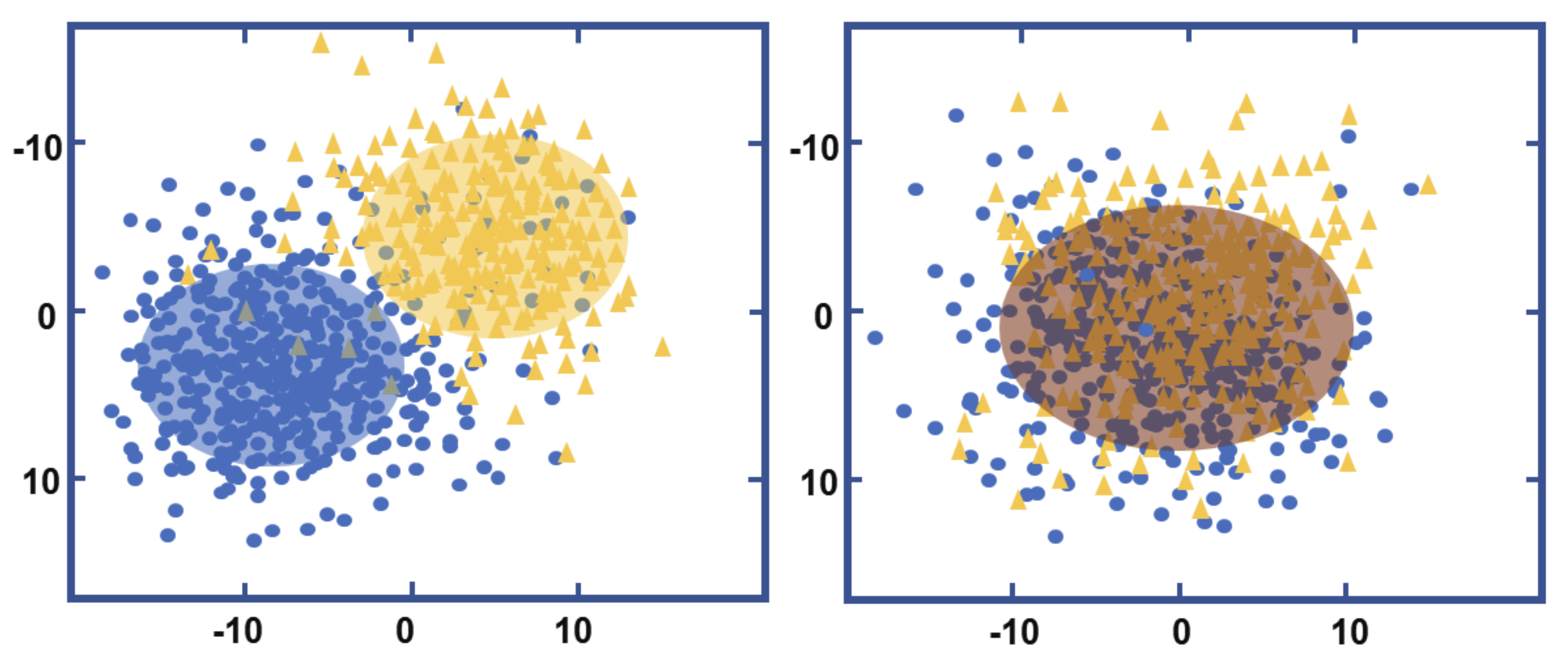}
	\caption{Visualization of target features in a sequence. (a) Projected features of the baseline tracker. (b) Projected features of our MArMOT with RT-MDNet as baseline. Herein, the blue circles and yellow triangles represent target features of RGB and NIR modalities respectively. From the results we can see that the heterogeneous gap of the target object between different modalities is mitigated to some extent.}
	\label{fig::scatter_map}
\end{figure}

To visually demonstrate the effectiveness of our method, we present the features of an example obtained by the baseline tracker RT-MDNet~\cite{jung2018real} and by our tracker MArMOT$\bf_{RT-MDNet}$ after being projected to the 2D space via the t-SNE algorithm ~\cite{van2008visualizing}, respectively, as shown in Fig.~\ref{fig::scatter_map}. It can be found that the gap between target features of RGB and NIR can be eliminated well after the introduction of the proposed algorithm.
%Moreover, this layer is trained using both modal data and can thus mitigate the appearance gap of target object between two modalities. %Fig.~\ref{fig::feature_visualization} shows an example to demonstrate the effectiveness of the modality-aware fusion layer. For efficiency, we implement it using a $1 \times 1$ convolution layer.

% {\flushleft \bf Domain-specific module}.
% Followed by RT-MDNet~\cite{jung2018real}, we design the domain-specific module to capture instance-aware feature representations. 
% To improve the efficiency, we first introduce the RoIAlign pooling layer to allow features of candidate regions be directly extracted on feature maps, which greatly accelerates feature extraction in tracking process.
% After that, three fully connected layers (fc4-6) are used to accommodate appearance changes of instances in different videos and frames. 
% Finally, we use the softmax cross-entropy loss and instance embedding loss to perform the binary classification to distinguish the foreground and background.

\begin{figure}[t]  %[]
	\centering
	%\begin{center} % 图片居中
	%  \fbox{\rule{0pt}{2in} \rule{.9\linewidth}{0pt}}
	\includegraphics[width=\linewidth,trim=0 0 0 0,clip]{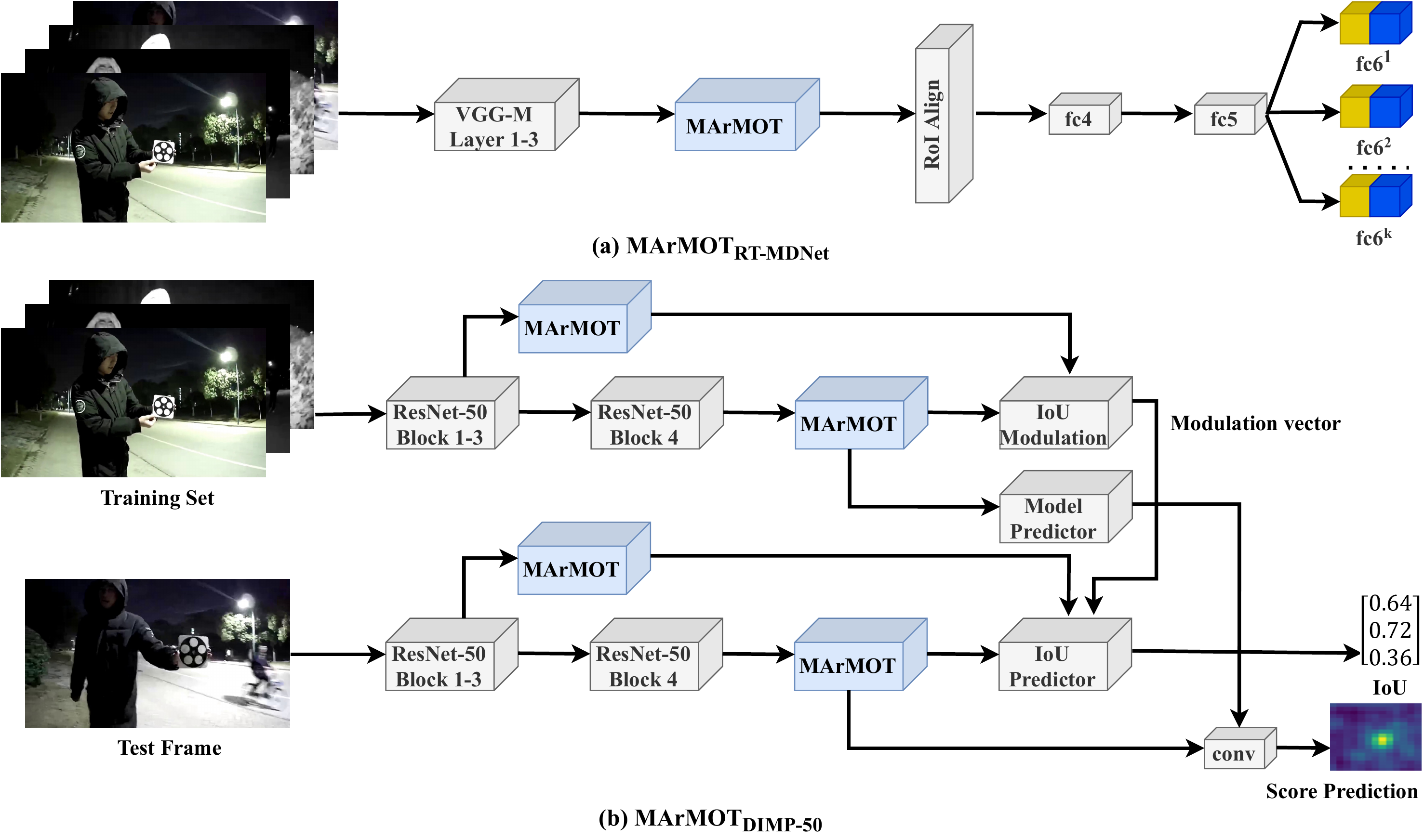}
	%\end{center}
	\caption{Visualization of the tracking architectures with MArMOT. (a) and (b) show the detailed structures of MArMOT combining with RT-MDNet and DiMP-50.}%, respectively. Better views in color with zoom-in.}
	%Bottom:Architecture of each component in MArMOT. The BN+ReLU layers after each Conv layer and the ReLU layers after each FC layer are omitted for clarity. Herein,$\oplus$ and $\otimes$ denote the operation of element-wise addition and multiplication respectively. F\_RGB and F\_NIR denote the output feature of RGB and NIR after two parallel modality-aware branches respectively. GAP is the representative of the global average pooling operation.}
	\label{fig::overall_network}
\end{figure}

\subsection{Tracking Architectures with MArMOT}

We embed the proposed plug-and-play MArMOT model into two tracking frameworks, i.e., RT-MDNet~\cite{jung2018real} and DiMP-$50$~\cite{bhat2019learning}, named MArMOT$\bf_{RT-MDNet}$ and MArMOT$\bf_{DiMP-50}$ respectively, to verify the effectiveness and generalization of MArMOT. The overall tracking frameworks are shown in Fig.~\ref{fig::overall_network}.

For each tracking framework, we first use the backbone network to extract deep feature representations of the target, then embed the proposed MArMOT model to mitigate the appearance gap of the target representations between different modalities, and finally send it to the classification branch and regression branch of target localization.
%MArMOT$\bf_{RT-MDNet}$ MArMOT$\bf_{DiMP-50}$
Specifically, on the tracking framework of RT-MDNet, we use the first three layers of VGG-M to capture modality-shared features of the target. Then, we insert our MArMOT model after the third layer to achieve modality-aware feature representation learning. More details are are shown in Fig.~\ref{fig::overall_network}(a). 
As for the DiMP-$50$ tracking framework, the input features of IoU predictor (regression branch) and model predictor (classification branch) are not the same layer.
Therefore, we insert the MArMOT model after the third and fourth blocks of ResNet$50$ for IoU predictor and model predictor.
For more details, the tracking framework is shown in Fig.~\ref{fig::overall_network}(b).
%we insert the MArMOT model after the third and fourth blocks of ResNet$50$ to meet the needs of IoU predictor module (regression branch) and model predictor module (classification branch) respectively, which is shown in Fig.~\ref{fig::overall_network}(b).

\begin{figure}[t]  %[]
	\centering
	%\begin{center} % 图片居中
	%  \fbox{\rule{0pt}{2in} \rule{.9\linewidth}{0pt}}
	\includegraphics[width=\linewidth,trim=0 0 0 0,clip]{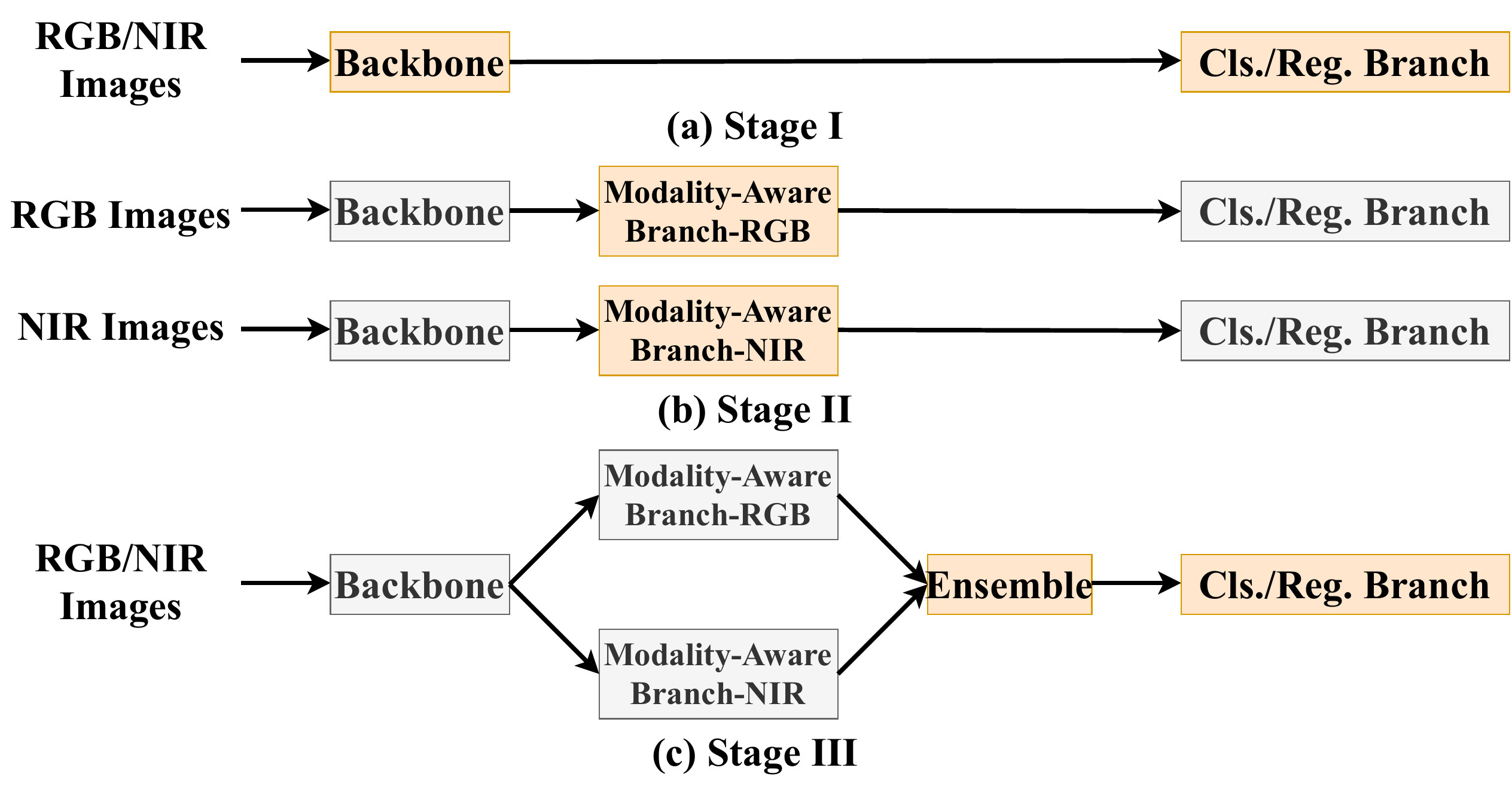}
	%\end{center}
	\caption{The visualization of three-stage training method. The parameters learned in each stage are shown in orange.}
	\label{fig::three-stage-training}
\end{figure}

\subsection{Three-stage Learning Algorithm}
There are two problems in training the entire tracking frameworks.
First, the loss of a training sample with any modality will be backwardly propagated to two modality-aware branches. Thus, there is no guarantee that the two modality-aware branches will learn the corresponding modality-specific representation of the target. Second, the modality information is available in training stage but unavailable in testing stage. Therefore, we need to train an ensemble layer to simulate the modal agnostic situation in tracking process.
To handle these two problems, we design an effective three-stage training algorithm.
\begin{itemize}
  \item Stage I : Fine-tune the parameters of the baseline network on our dataset. Note that our dataset is the first cross-modal tracking dataset.
  To adapt the tracker to the cross-modal scenario, we first need to fine-tune the parameters of the baseline network pre-trained on other large-scale datasets on our training set. The learning rate of the network parameters is set to one-tenth of the default learning rate of the baseline network, and the number of iterations remains the same. %We select the fine-tuned tracker parameters as the model of the baseline tracker.
  \item Stage II : Train two parallel modality-aware branches. To enable two parallel modality-aware branches to learn the modality-specific representations of the target in different modalities, we first divide the training set into two subsets according to modality type, and use the corresponding sub-datasets to learn the parameters of the corresponding modality-aware branches. 
  Since the baseline network has been adapted to the cross-modal tracking task in the first stage, thus, in this stage, we only learn the parameters of the two modality-aware branches, and the rest of the parameters are fixed (except for the parameters of the fc$6$ layer based on the RT-MDNet~\cite{jung2018real} framework).
  The initial learning rate is set to $1e-6$ and $1e-4$, and the number of iterations is set to $50$ and $1000$, respectively.
  \item Stage III : Train ensemble layer and fine-tune the parameters of the baseline network on our dataset again. After the first two stages of training, the baseline network has been able to adapt to the tracking of cross-modal scenarios, and the two parallel modality-aware branches have also learned the modality-specific representations of the target in different modalities. 
  Since which modality in each frame is unknown in the tracking process, the deep features extracted by the backbone need to be sent to two parallel modality-aware branches to extract the corresponding modality-specific representations. To simulate the modality-unknowable situation in the tracking process, we train the ensemble layer at this stage to perform weighted fusion of the features of the two branches, and fine-tune the parameters of the network to adapt to the situation after embedding the proposed MArMOT. At this stage, we only learn the parameters of the ensemble layer and fine-tune the parameters of the baseline network except the backbone. The learning rate of the ensemble layer is the same as the modality-aware branch of the second stage, and the learning rate of the tracker is the same as the first stage, and the number of iterations is set to the same as the second stage.
\end{itemize}
Fig.~\ref{fig::three-stage-training} shows more details, and the learned part is indicated in the orange color.

\subsection{Online Tracking}
The tracking processes and parameter settings of our trackers during online tracking are the most same as the baseline trackers.
%The online tracking process and settings are consistent with the baseline algorithms
The only difference is that the deep features extracted by the backbone networks (VGG-M in MArMOT$\bf_{RT-MDNet}$ and ResNet$50$ in MArMOT$\bf_{DiMP-50}$) are sent to the proposed MArMOT model to mitigate the appearance gap of the target representations under different modalities.
The outputs of MArMOT are as the inputs of the classifiers (fc$4-$fc$6$ in MArMOT$\bf_{RT-MDNet}$ and model predictor in MArMOT$\bf_{DiMP-50}$) and the regressor (IoU predictor module in MArMOT$\bf_{DiMP-50}$).
The details of tracking processes of our trackers are shown in Fig.~\ref{fig::overall_network}.

\section{CMOTB Benchmark}
Large-scale dataset are crucial in cross-modal object tracking because they are not only useful for training deep trackers, but also for evaluating different tracking algorithms. To this end, we provide a large-scale cross-modal object tracking benchmark, called CMOTB.
In this section, we introduce CMOTB with detailed analysis. 

\subsection{Data Collection and Annotation}
{\flushleft \bf Large-scale collection.}
Current object tracking field lacks cross-modal video data, and we introduce our CMOTB benchmark. Our goal is to provide large-scale and high-diverse cross-modal object tracking benchmark for real-world scenarios and challenges. 
To this end, we use hand-held cameras to capture video data in a large range of scenes and background complexities. Unlike traditional visual tracking data, we need to consider the variations of light intensity that trigger modality switch in data creation. Therefore, we carefully select some environmental conditions to simulate real-world applications such as visual surveillance, intelligent transportation and self-driving systems. 
Fig.~\ref{fig::motivation} shows a typical example of CMOTB dataset, and we can see that the imaging is switching with several times between RGB and NIR modalities.
By this way, we collect $654$ cross-modal image sequences with over $481$K frames in total and the average video length of is more than $735$ frames.

\begin{table}[t]
\centering
	\caption{The details of our CMOTB Dataset}\small
	\resizebox{.4\textwidth}{!}{\begin{tabular}{c|ccccc}
		\hline
		Benchmark & Video & \begin{tabular}[c]{@{}c@{}}Min \\ frames\end{tabular} & \begin{tabular}[c]{@{}c@{}}Mean\\ frames\end{tabular} & \begin{tabular}[c]{@{}c@{}}Max \\ frames\end{tabular} & \begin{tabular}[c]{@{}c@{}}Total \\ frames\end{tabular}   \\
		\hline
		\textbf{CMOTB (train)} & \textbf{438} & \textbf{85} & \textbf{750} & \textbf{2037} & \textbf{328K} \\
		\textbf{CMOTB (test)} & \textbf{216} & \textbf{101} & \textbf{712} & \textbf{1838} & \textbf{153K}\\
		\hline
	\end{tabular}}
	\label{tb::information}
\end{table}

We present the detailed information of CMOTB in Table~\ref{tb::information}. Note that there is no other cross-modal object tracking dataset, therefore, we divide CMOTB into training set and testing set for facilitating the training of deep trackers for cross-modal object tracking.

{\flushleft \bf High-quality dense annotation.}
We use a minimum bounding box to represent object states including position and scale, and annotate each frame for training and evaluation set. Since the labeling process is time-consuming and labor-intensive, we design an auxiliary labeling tool based on the ViTBAT~\cite{biresaw2016vitbat}. The tool allows the labeling of their states manually or semi-automatically through a simple and friendly user interface in a time-efficient manner. 
The generated bounding boxes are accurate in most situations. However, when the object undergoes drastic appearance variations, the generated bounding boxes might be not quite accurate. For these bounding boxes, we manually adjust them carefully. 

To ensure high-quality annotations, we train $4$ professional annotators to learn consistent annotation standards. Moreover, we let professional checkers to carry out a frame-by-frame inspection to prevent wrong and inaccurate marking. Due to the special challenges brought by modality switch, some objects are sometimes temporarily invisible, which may result in losing a few frames or more than a dozen frames. For such scenarios, we will keep ground truths unchanged of target object until it is visible.

\begin{table}[!tp]
	\centering
	\caption{Distribution of the number of modality switch.}
	\resizebox{.475\textwidth}{!}{
		\begin{tabular}{c|ccccc}
			\hline
			Modality switch & Once & Twice & Three times &  More than three times \\
			\hline
			Number of sequences & \textbf{410}  & \textbf{182}  & \textbf{47}  & \textbf{15} \\
			\hline               
		\end{tabular}%
	}
	\label{tb::modalityswitch}
\end{table}

\begin{table}[t]
\centering
	\caption{Descriptions of 11 different attributes in CMOTB.}
	\resizebox{.45\textwidth}{!}{\begin{tabular}{p{30pt}p{300 pt}}
		%\toprule
		\hline
		\textbf{Attribute}& \textbf{Definition}\\
		\hline
		\textbf{SV }& Scale Variation - The ratio of bounding box is outside the rage $[0.5, 2]$.\\
		\textbf{BC}& Background Clutter - The background has the similar appearance as the target.\\
		\textbf{ARC}& Aspect Ratio Change - The ratio of bounding box aspect ratio is outside the rage $[0.5, 2]$.\\
		\textbf{SO}& Similar Object - There are objects of similar shape or same type near the target.\\
		\textbf{FM}& Fast Motion - The motion of the target is larger than the size of its bounding box.\\
		\textbf{IPR}& In-Plane Rotation - The target rotates out of the image plane.\\
		\textbf{OV}&Out-of-View - The target completely leaves the video frame.\\
		\textbf{PO}&Partial Occlusion - The target is partially occluded.\\
		\textbf{MA}& Modality Adaptation -  Frame content has high intensity due to imaging adaptation to the environment in switching.\\
		\textbf{FO}& Full Occlusion - The target is fully occluded.\\
		\textbf{MB}& Motion Blurred - The target region is blurred due to target or camera motion.\\
		%\bottomrule
		\hline
	\end{tabular}}
	\label{tb::attribute}
\end{table}

% \begin{figure}[!htp]  %[!htp]
% 	\centering
% 	%\begin{center} % 图片居中
% 	%  \fbox{\rule{0pt}{2in} \rule{.9\linewidth}{0pt}}
% 	\includegraphics[width=.75\linewidth]{figure/attribute.png}
% 	%\end{center}
% 	\caption{Distribution of attribute-based sequences on CMOTB testing set.}
% 	\label{fig::attribute}
% \end{figure}

\begin{table}[t]
	\centering
	\caption{Distribution of attribute-based sequences on CMOTB testing set.}
	\resizebox{0.475\textwidth}{!}{%
		\begin{tabular}{cccccccccccc}
			\hline
			Attribute & SV & BC & ARC & SO & FM & IPR & OV & PO & MA & FO & MB  \\
			\hline
			Number    & \textbf{38} & \textbf{61} & \textbf{23} & \textbf{72} & \textbf{27} & \textbf{134} & \textbf{31} & \textbf{115} & \textbf{97} & \textbf{47} & \textbf{53}\\
			\hline              
		\end{tabular}%
	}
	\centering
	\label{tb::attribute2}
\end{table}

\subsection{Attributes}
Existing multi-modal tracking datasets, e.g., RGBD ~\cite{Song13iccv} and RGBT ~\cite{li2016learning,li2019rgb}, include two-modal data in each frame, while our dataset has only one modality in each frame but might occur modality switch. This is the major difference from existing multi-modal tracking datasets.
The modality switch means that the imaging is changed from one modality to another one caused by light intensity variation. In such scenarios, the appearance of target object usually varies significantly so that trackers are easily failed. 
Note that the number of modality switch in a sequence is a key factor in affecting trackers. Therefore, we take the switch times in data creation and report the data distribution on switch times in Table~\ref{tb::modalityswitch}.

According the modality switch, a new attribute, i.e., modality adaptation, is thus introduced in CMOTB.
The modality adaptation means that some frames have high intensity due to imaging adaptation to the environment in modality switch. It does not always occur when imaging is switch, and thus we take it as an attribute.
% Fig.~\ref{fig::attribute} 
To enable attribute-based performance analysis of trackers, we annotate each sequence with several attributes from the total $11$ attributes, including Scale Variation (SV), Aspect Ratio Change (ARC), Fast Motion (FM), Out-of-View (OV), Modality Adaptation (MA), Motion Blur (MB), Background Clutter (BC), Similar Object(SO), In-Plane Rotation (IPR), Partial Occlusion (PO) and Full Occlusion (FO). The attributions are defined in Table \ref{tb::attribute}, and Table \ref{tb::attribute2} shows the video distribution of attributes on the testing set.

\subsection{Statistics}

%#########  attribute balancing 
CMOTB consists of $654$ video sequences, which cover most of the challenges in real-world scenarios. According to the holdout method~\cite{kohavi1995study}, we randomly split the testing and training sets of our dataset with the ratio of $1:2$. And We have counted the distribution of attributes on testing set in Table \ref{tb::attribute2}. The total number of frames of CMOTB reaches $481$K, and the average length of our video sequence and the maximum number of frames reach $735$ and $2037$ frames respectively. More details are shown in Table~\ref{tb::information}.
% \begin{figure}[]  %[!htp]
% 	\centering
% 	\includegraphics[width=\linewidth]{figure/feature.eps}
% 	\caption{Illustration of feature maps in the challenge of modality switch. (a) Input two adjacent frames, where the modality switch occurs. (b) Feature map visualization of the baseline tracker, RT-MDNet~\cite{jung2018real}. (c) Feature map visualization of our MArMOT. From the results one can see that our MArMOT can adapt well in modality switch. }
% 	\label{fig::feature_visualization}
% \end{figure}

\subsection{Discussion}

{\bf \flushleft Differences from relevant tasks}.
We discuss the differences of our new task from the task of multi-modal visual object tracking. Existing work usually introduce thermal infrared or depth information to achieve multi-modal visual object tracking, called RGBT tracking ~\cite{li2019rgb,li2020challenge} and RGBD tracking ~\cite{Song13iccv}. 
Comparing with multi-modal visual object tracking, our task has the following differences and advantages. First, our task is more practical. Many visual cameras have equipped with NIR imaging, but RGBT or RGBD tracking requires two cameras. Second, our task is more cost-effective. Thermal cameras are usually very expensive and depth sensors have limited imaging range and environment, but our task only relies on surveillance cameras and thus does not have these limitations. Finally, the multi-modal data in our task do not have any alignment error. Both RGBT and RGBD tracking tasks involve two cameras and the alignment cross different modalities is needed, while our imaging system only includes one camera whose imaging is switchable between RGB and NIR modalities.

\section{Experiment}
\subsection{Evaluated algorithms}
%To demonstrate the effectiveness of the proposed method, we select $17$ deep learning based trackers.
We evaluate $17$ most advanced and representative trackers on our benchmark. These trackers cover mainstream tracking algorithms from $2016$ to $2020$, and they are MDNet~\cite{nam2016learning}, RT-MDNet~\cite{jung2018real}, SiamFC~\cite{bertinetto2016fully}, SPLT~\cite{yan2019skimming}, GradNet~\cite{li2019gradnet}, TACT~\cite{choi2020visual}, SiamMask~\cite{wang2019fast}, VITAL~\cite{song2018vital}, GlobalTrack~\cite{huang2020globaltrack}, SiamRPN++~\cite{li2019siamrpn++}, ATOM~\cite{danelljan2019atom}, DiMP-$50$\cite{bhat2019learning}, SiamBAN~\cite{chen2020siamese}, SiamDW~\cite{zhang2019deeper}, LTMU~\cite{dai2020high}, Ocean~\cite{zhang2020ocean} and DaSiamRPN~\cite{zhu2018distractor}. %Table~\ref{tab::summary_trackers} summarizes the selected trackers.
It should be noted that DiMP-$50$ and RT-MDNet are two representative tracking frameworks based on regression and classification, respectively. Thus, to demonstrate the effectiveness of the proposed method, we embed MArMOT into these two frameworks, named MArMOT$\bf_{DiMP-50}$ and MArMOT$\bf_{RT-MDNet}$, respectively. It is worth noting that all algorithms are evaluated on our testing set using the model provided by authors. %In order to verify the effectiveness of our method, we also provide the fine-tuned results of the DiMP-$50$ and RT-MDNet algorithms on our training set as baseline, denoted as DiMP-$50$* and RT-MDNet* respectively.

\subsection{Evaluation metrics}
To evaluate the performance of different trackers, we employ the widely used tracking evaluation metrics including precision rate (PR), normalized precision rate (NPR) and success rate (SR) ~\cite{muller2018trackingnet} for quantitative performance evaluation. 
PR is the percentage of frames whose distance of the estimated bounding boxes with the ground truth is below a predefined threshold, to rank the trackers, we set the distance threshold to $20$ pixels to compute the representative PR. 
However, since the PR is very sensitive to the target size, thus, we normalize the PR on the size of the ground truth to calculate the normalized precision rate, distance threshold is also set to $20$ pixels to compute the representative NPR.
SR is the percentage of frames where the overlap rate between the estimated bounding boxes and the ground truth are greater than a threshold, we set the overlap ratio to $0.5$ and use the area under curve of SR plots to compute two kinds of representative SR scores respectively, denoting SR-I score and SR-II score for the clarity.

\begin{figure*}[!htp]
	\centering
	\subfigure[]{
		\begin{minipage}[b]{0.31\linewidth}
			\includegraphics[width=\linewidth,trim=40 0 40 40,clip]{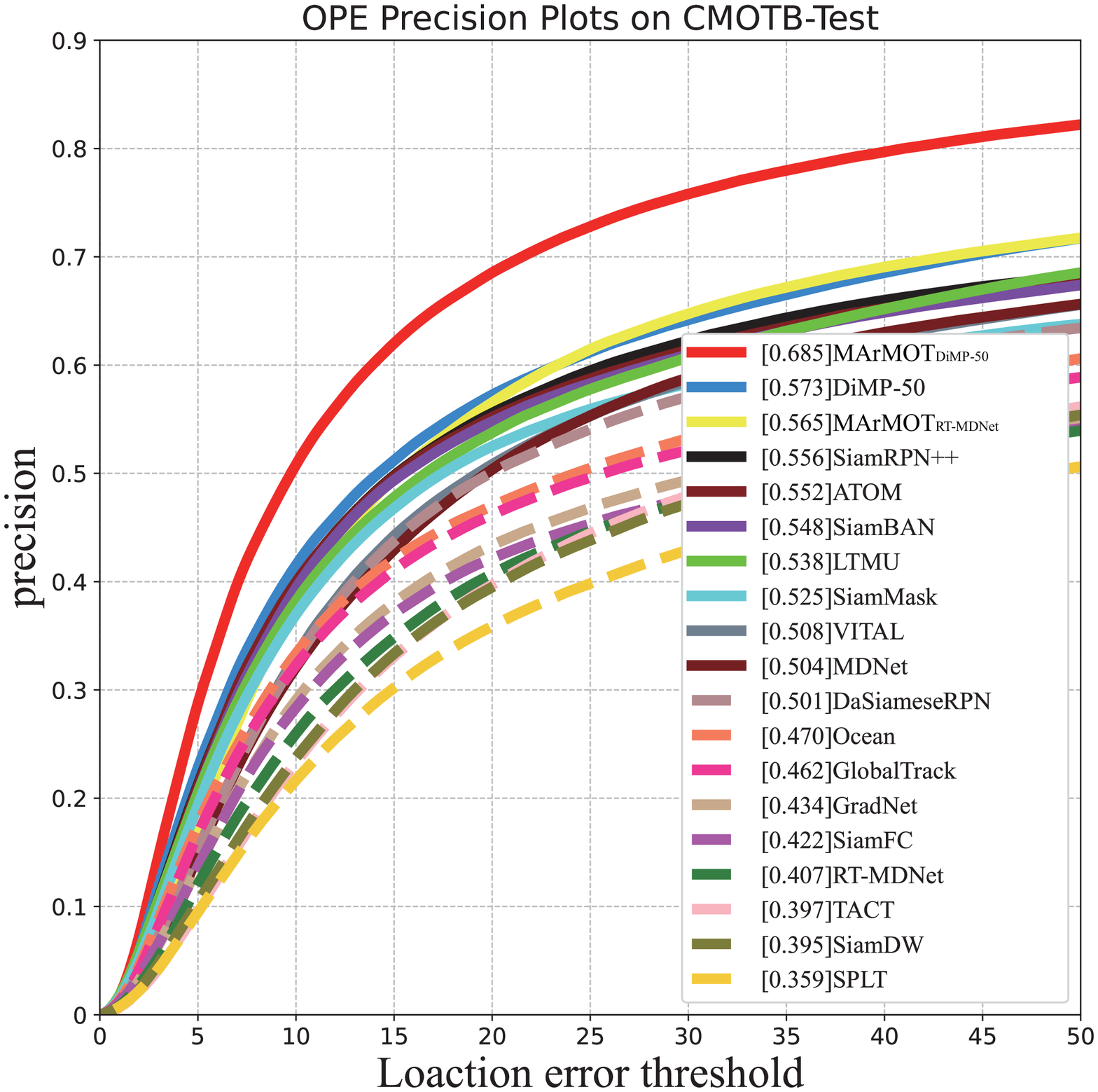}
			%\caption{}
		\end{minipage}%
	}%
	\subfigure[]{
		\begin{minipage}[b]{0.31\linewidth}
			\includegraphics[width=\linewidth,trim=40 0 40 40,clip]{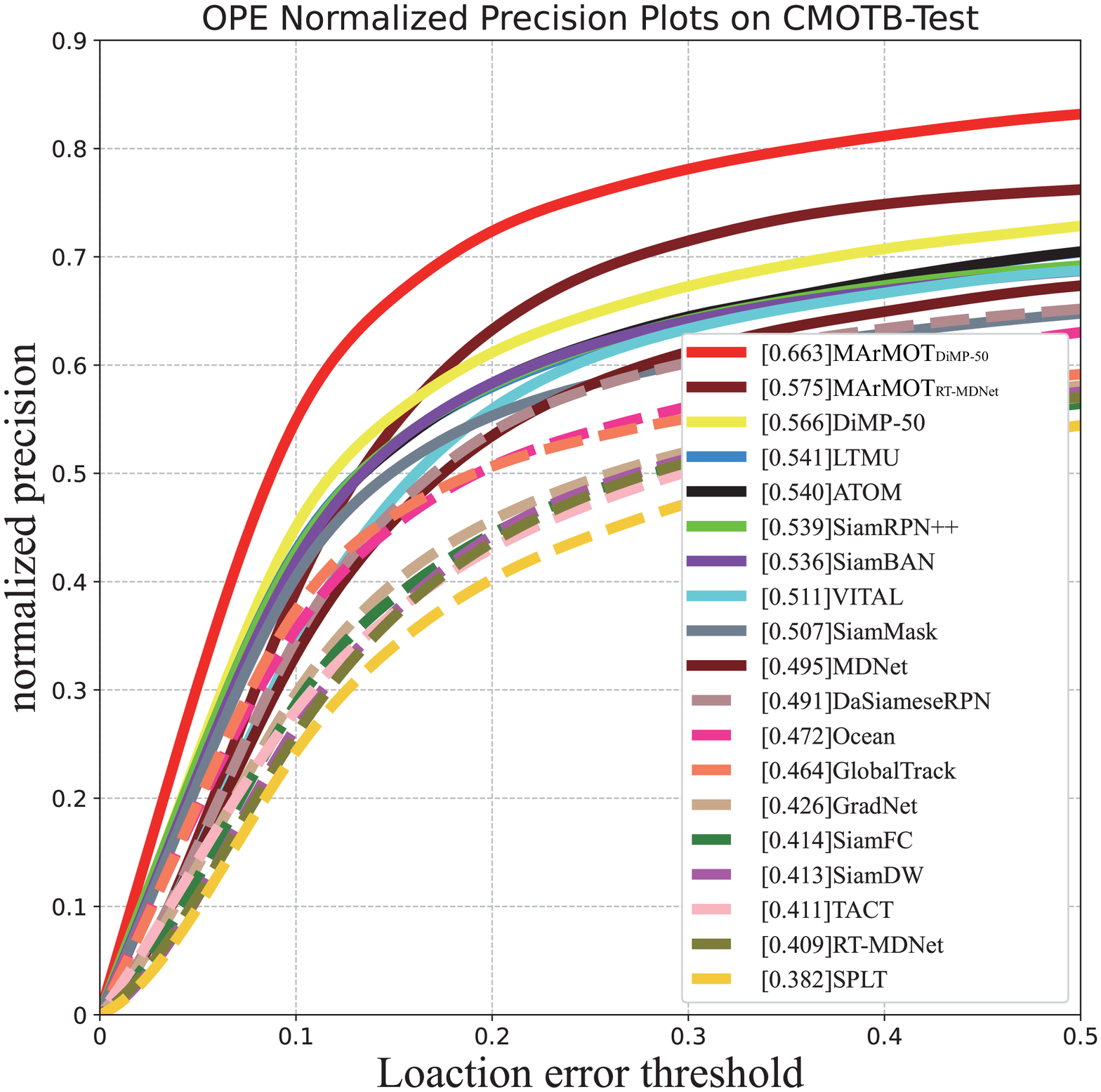}
			%\caption{fig2}
		\end{minipage}%
	}%
	\subfigure[]{
		\begin{minipage}[b]{0.31\linewidth}
			\includegraphics[width=\linewidth,trim=40 0 40 40,clip]{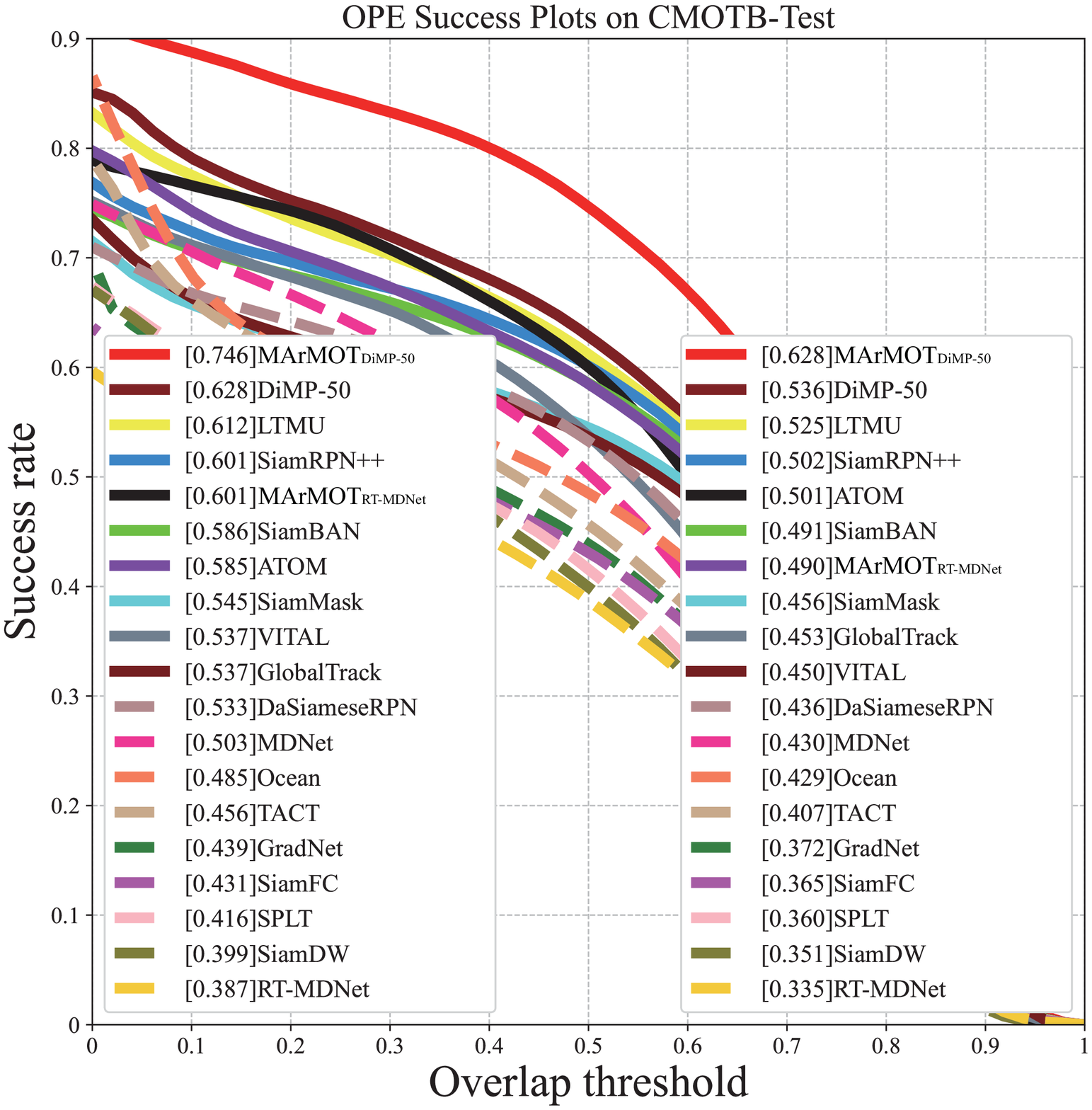}
			%\caption{fig2}
		\end{minipage}%
	}%
	\caption{ Tracking curves of $17$ trackers on the CMOTB testing dataset. In (c), we show two kinds of representative SR scores in the legends, and the left and right ones are SR-I and SR-II respectively. }
	\label{fig::tracking_result}
\end{figure*}

\subsection{Overall Performance}
We present the tracking performance in terms of precision plots, normalized precision plots and success plots in Fig.~\ref{fig::tracking_result}, and the representative scores are shown in the legends. 

{\bf Regression based deep trackers}. 
Regression-based trackers such as DiMP-$50$, LTMU, SiamRPN++, SiamBAN, ATOM, SiamMask, have achieved high performance while running at real-time speed.
They are usually offline trained to learn a powerful regressor from large-scale datasets to locate the target. However, their performance is limited in cross-modal object tracking due to the existence of large heterogeneous gap between RGB and NIR modalities, as shown in Fig.~\ref{fig::tracking_result}.
%Take the best-performing DiMP-$50$ as an example. After fine-tuning on our training set, DiMP-$50$* advance the untuned version DiMP-$50$ with $6.9\%$/$6.4\%$/$8.2\%$/$6.3\%$ gains in PR/NPR/SR-I/SR-II, which prove the necessity of our proposed dataset to solve the cross-modal object tracking task.
To verify the effectiveness of the proposed method, we insert the proposed MArMOT into the DiMP-$50$ tracking framework, namely MArMOT$\bf_{DiMP-50}$, and train the entire framework through the proposed multi-stage training method. Our MArMOT$\bf_{DiMP-50}$ outperforms the baseline tracker DiMP-$50$ with $11.2\%$/$9.7\%$/$11.8\%$/$9.2\%$ gains in PR/NPR/SR-I/SR-II, and has excellent performance gains compared with all comparison methods.

{\bf Classification based deep trackers}.
Classification based deep trackers such as MDNet, RT-MDNet and VITAL, usually employ online learning to train binary classifiers using positive and negative samples, and thus have a good generalization ability. %Because of their strong generalization ability and good robustness, they can adapt to the appearance change of the target well, and have a certain anti-jamming ability when facing the challenge of modality switch. 
%In order to verify the necessity of the proposed dataset, we test the model provided by the author(namely RT-MDNet) and the model fine-tuned on our training set (namely RT-MDNet*) on the RT-MDNet tracking framework, respectively.
%We can find from the Fig.~\ref{fig::tracking_result} that compared with untuned version RT-MDNet, RT-MDNet* advance with $11.0\%$/$11.4\%$/$14.7\%$/$10.9\%$ gains in PR/NPR/SR-I/SR-II.
To verify the effectiveness and generalization of the proposed method, we also insert MArMOT into the RT-MDNet framework, namely MArMOT$\bf_{RT-MDNet}$, and train the entire framework through the proposed multi-stage training method. And our MArMOT$\bf_{RT-MDNet}$ outperforms the baseline tracker RT-MDNet with $15.8\%$/$16.6\%$/$21.4\%$/$15.5\%$ gains in PR/NPR/SR-I/SR-II.
We can find that although the performance of the baseline tracker is very low, but it surpasses the performance of all classification-based tracking frameworks after introducing our proposed model, which proves the effectiveness of our proposed method.

%These results fully validate the effectiveness of our MArMOT. Meanwhile our pre-trained data are much less than ATOM and LTMU, although our method is slight worse than them in PR and SR-II. In specific, both ATOM and LTMU use the pre-trained model on ImageNet, COCO, LaSOT and TrackingNet, and we only use ImageNet. Second, we do not use some advanced techniques, such as IOU-Net, model verifier and re-detector, which are used in some trackers (e.g.,ATOM and LTMU). Finally, we use simple VGG as backbone network, while ATOM and LTMU use ResNet.

% \begin{table}[t]
% 	\centering
% 	\caption{Speed comparisons of different trackers without and with MArMOT.}
% 	\resizebox{0.475\textwidth}{!}{%
% 		\begin{tabular}{ccccc}
% 			\hline
% 			Trackers & RT-MDNet & MArMOT$\bf_{RT-MDNet}$ & DiMP-$50$ & MArMOT$\bf_{DiMP-50}$  \\
% 			\hline
% 			FPS      & \bf{29}         & \bf{24}  &   \bf{32}     & \bf{25}\\
% 			\hline              
% 		\end{tabular}%
% 	}
% 	\centering
% 	\label{tb::Runtime analysis}
% \end{table}

\begin{table*}[t]
	\centering
	\caption{SR-I scores of different trackers on 11 challenging attributes using the CMOTB testing sets. And speed comparisons of different trackers. Herein, the best and second best results are indicated by \textcolor{red}{red} and \textcolor{blue}{blue} fonts.}
	{%
	\resizebox{.7\textwidth}{!}{
		\begin{tabular}{c|ccccccccccc|c}
			\hline
			\textbf{Trackers} & SV & BC & ARC & SO & FM & IPR & OV & PO & MA & FO & MB & FPS \\
			\hline
			\textbf{MArMOT$\bf_{DiMP-50}$} & \textcolor{red} {0.806} & \textcolor{red}{0.708} & \textcolor{red}{0.788} & \textcolor{red}{0.688} & \textcolor{red}{0.752} & \textcolor{red}{0.727} & \textcolor{red}{0.831} & \textcolor{red}{0.702} & \textcolor{red}{0.749} & \textcolor{red}{0.697} & \textcolor{red}{0.654} & 25\\
% 			\textbf{DiMP-50*} & \textcolor{blue}{0.759} & \textcolor{blue}{0.682} & \textcolor{blue}{0.755} & \textcolor{blue}{0.625} & \textcolor{blue}{0.720} & \textcolor{blue}{0.682} & \textcolor{blue}{0.807} & \textcolor{blue}{0.679} & \textcolor{blue}{0.710} & \textcolor{blue}{0.682} & \textcolor{red}{0.656} & \textcolor{blue}{0.710}\\
			\textbf{DiMP-50} & 0.693 & \textcolor{blue} {0.610} & \textcolor{blue} {0.663} & 0.561 & \textcolor{blue} {0.616} & 0.599 & 0.717 & \textcolor{blue} {0.580} & 0.616 & \textcolor{blue} {0.597} & 0.552 & 32\\
			\textbf{LTMU} & \textcolor{blue} {0.710} & 0.601 & 0.658 & 0.582 & 0.517 & \textcolor{blue} {0.601} & \textcolor{blue} {0.742} & 0.569 & \textcolor{blue} {0.633} & 0.584 & 0.511 & 3\\
			\textbf{SiamRPN++} & 0.634 & 0.568 & 0.642 & \textcolor{blue} {0.602} & 0.525 & 0.587 & 0.652 & 0.569 & 0.597 & 0.530 & 0.507 & 21\\
			\textbf{MArMOT$\bf_{RT-MDNet}$} & 0.617 & 0.576 & 0.521 & 0.567 & 0.502 & 0.557 & 0.680 & 0.552 & 0.596 & 0.448 & \textcolor{blue} {0.576} & 24\\
			\textbf{SiamBAN} & 0.626 & 0.587 & 0.626 & 0.564 & 0.505 & 0.551 & 0.631 & 0.521 & 0.563 & 0.505 & 0.504 & 29\\
			\textbf{ATOM} & 0.628 & 0.526 & 0.650 & 0.525 & 0.550 & 0.577 & 0.633 & 0.523 & 0.583 & 0.508 & 0.470 & 28\\
			\textbf{SiamMask} & 0.591 & 0.538 & 0.613 & 0.533 & 0.550 & 0.522 & 0.612 & 0.486 & 0.542 & 0.457 & 0.438 & 41\\
			\textbf{VITAL} & 0.600 & 0.525 & 0.515 & 0.522 & 0.487 & 0.508 & 0.585 & 0.499 & 0.505 & 0.474 & 0.456 & 0.3 \\
			\textbf{GlobalTrack} & 0.506 & 0.531 & 0.559 & 0.458 & 0.427 & 0.515 & 0.602 & 0.482 & 0.558 & 0.533 & 0.496 & 1\\
            % \textbf{RT-MDNet*} & 0.591 & 0.491 & 0.514 & 0.518 & 0.516 & 0.485 & 0.603 & 0.491 & 0.512 & 0.490 & 0.560 & 0.534\\	
			\textbf{DasiamRPN} & 0.577 & 0.522 & 0.549 & 0.538 & 0.540 & 0.511 & 0.594 & 0.489 & 0.507 & 0.482 & 0.473 & 144\\
			\textbf{MDNet} & 0.531 & 0.477 & 0.494 & 0.480 & 0.457 & 0.473 & 0.540 & 0.460 & 0.457 & 0.468 & 0.457 & 1 \\
			\textbf{Ocean} & 0.518 & 0.444 & 0.578 & 0.446 & 0.502 & 0.460 & 0.539 & 0.448 & 0.480 & 0.452 & 0.424 & 42 \\
			\textbf{TACT} & 0.457 & 0.451 & 0.531 & 0.377 & 0.344 & 0.438 & 0.559 & 0.428 & 0.497 & 0.411 & 0.372 & 29\\
			\textbf{GradNet} & 0.476 & 0.430 & 0.424 & 0.429 & 0.424 & 0.437 & 0.441 & 0.407 & 0.419 & 0.384 & 0.375 & 68\\
			\textbf{SiamFC} & 0.524 & 0.376 & 0.376 & 0.448 & 0.473 & 0.410 & 0.487 & 0.405 & 0.377 & 0.377 & 0.394 & 44 \\	
			\textbf{SPLT} & 0.442 & 0.387 & 0.498 & 0.421 & 0.390 & 0.419 & 0.506 & 0.387 & 0.431 & 0.411 & 0.361 & 13\\
			\textbf{SiamDW} & 0.326 & 0.397 & 0.456 & 0.376 & 0.354 & 0.372 & 0.481 & 0.374 & 0.394 & 0.314 & 0.392 & 22\\
			\textbf{RT-MDNet} & 0.386 & 0.347 & 0.400 & 0.415 & 0.476 & 0.365 & 0.392 & 0.351 & 0.336 & 0.362 & 0.354 & 29\\
			\hline                 
		\end{tabular}%
		}
	}
	\centering
	\label{tb::results_challenges}
\end{table*}

{\bf Runtime analysis}.
To verify the influence of the proposed method on the tracking efficiency, we carried out an efficiency analysis of the tracker without and with the proposed MArMOT in DiMP-$50$ and RT-MDNet tracking frameworks. The experiments are run on two platforms with Intel(R) Xeon(R) Silver $4210$ CPU ($32$G RAM), GeForce RTX $3090$ GPU and Intel(R) Xeon(R) Silver $4210$ CPU ($32$G RAM), GeForce RTX $1080$Ti GPU for DiMP-$50$ and RT-MDNet respectively. The last column of Table ~\ref{tb::results_challenges} reports the running speeds of these trackers, which demonstrates that the tracking speeds decrease slightly with an additional MArMOT.

\subsection{Attribute-based Performance}
To analyze performance under different challenges faced by existing trackers, we evaluate our algorithm against $17$ tracking algorithms on $11$ attributes, as shown in Table~\ref{tb::results_challenges}. 
It can be seen from the table that our method MArMOT$\bf_{DiMP-50}$ achieves the best results in all attributes compared with all other algorithms.
In addition, after adding our module to the two frameworks, it can also be concluded that every attribute has an excellent performance improvement compared with the baseline method, which proves the effectiveness and generalization ability of our method.

\begin{table}[t]
	\centering
	\caption{ Comparison of deep trackers re-trained on CMOTB dataset, where * indicates the tracker is re-trained using CMOTB training dataset, and $\uparrow$ indicates the performance improvements with re-training over their own baselines.}
	\resizebox{.45\textwidth}{!}{%
		\begin{tabular}{c|cccc}
			\hline
			Trackers & PR & NPR  & SR-I & SR-II \\
			\hline
			\textbf{DiMP-50}* & 0.642$^{\uparrow 6.9\%}$ & 0.630$^{\uparrow 6.4\%}$  & 0.710$^{\uparrow 8.2\%}$ & 0.599$^{\uparrow 6.3\%}$\\
			\textbf{RT-MDNet}* & 0.517$^{\uparrow 11.0\%}$ & 0.523$^{\uparrow 11.4\%}$  & 0.534$^{\uparrow 14.7\%}$ & 0.444$^{\uparrow 10.9\%}$\\
			\textbf{LTMU}* & 0.561$^{\uparrow 2.3\%}$ & 0.563$^{\uparrow 2.2\%}$  & 0.628$^{\uparrow 1.6\%}$ & 0.526$^{\uparrow 0.1\%}$\\
			%\textbf{SiamBAN}* & 0.556$^{\uparrow 0.8\%}$ & 0.541$^{\uparrow 0.5\%}$  & 0.601$^{\uparrow 1.5\%}$ & 0.493$^{\uparrow 0.2\%}$\\
			\textbf{SiamRPN++}* & 0.574$^{\uparrow 1.8\%}$ & 0.552$^{\uparrow 1.3\%}$  & 0.612$^{\uparrow 1.1\%}$ & 0.503$^{\uparrow 0.1\%}$\\
			%\textbf{VITAL}* & 0.526$^{\uparrow 1.8\%}$ & 0.539$^{\uparrow 2.8\%}$  & 0.551$^{\uparrow 1.4\%}$ & 0.483$^{\uparrow 3.3\%}$\\
			\textbf{SiamMask}* & 0.556$^{\uparrow 3.2\%}$ & 0.529$^{\uparrow 2.2\%}$  & 0.571$^{\uparrow 2.6\%}$ & 0.486$^{\uparrow 3.0\%}$\\
			\textbf{MDNet}* & 0.534$^{\uparrow 3.9\%}$ & 0.526$^{\uparrow 3.1\%}$  & 0.547$^{\uparrow 4.4\%}$ & 0.461$^{\uparrow 3.1\%}$\\
			\textbf{GlobalTrack}* & 0.495$^{\uparrow 3.1\%}$ & 0.482$^{\uparrow 2.0\%}$  & 0.555$^{\uparrow 1.8\%}$ & 0.477$^{\uparrow 2.4\%}$\\
			\hline
			\textbf{MArMOT$\bf_{DiMP-50}$} & 0.685 & 0.663  & 0.746 & 0.628\\
			\textbf{MArMOT$\bf_{RT-MDNet}$} & 0.565 & 0.575  & 0.601 & 0.490 \\  
			\hline              
		\end{tabular}%
	}
	\centering
	\label{tb::retrain}
\end{table}

\subsection{Training Dataset Validation}
We select seven representative trackers including DiMP-$50$, RT-MDNet, LTMU, SiamRPN++, SiamMask, MDNet and GlobalTrack to demonstrate the effectiveness of our training dataset in the training of deep models. The results are shown in Table~\ref{tb::retrain}, which shows that all the re-trained deep trackers achieve obvious improvements and verify the necessity of proposing this dataset for the study of cross-modal object tracking. %Note that our MArMOT still performs favorably against these re-trained trackers, suggesting the effectiveness of our approach.
In addition, after adding our proposed model to the DiMP-$50$ and RT-MDNet frameworks, we can see that the performance has been further improved, i.e., $4.3\%$/$3.3\%$/$3.6\%$/$2.9\%$ and $4.8\%$/$5.2\%$/$6.7\%$/$4.6\%$ gains on the PR/NPR/SR-I/SR-II respectively, which proves the effectiveness of MArMOT.

\subsection{Analysis of MArMOT}

\begin{table}[htp]
	\centering
	\caption{Tracking results of trackers on the synthetic RGBT dataset, where * indicates the tracker is re-trained using synthetic RGBT training dataset.}
	\resizebox{.5\textwidth}{!}{%
		\begin{tabular}{c|cccc}
			\hline
			Trackers & PR & NPR & SR-I & SR-II  \\
			\hline
			\textbf{DiMP-50-RGBT} & 0.681 & 0.468  & 0.523 & 0.446\\
			\textbf{DiMP-50*-RGBT} & 0.757 & 0.507 & 0.602 & 0.502\\
			\hline   
			\textbf{MArMOT$\bf_{DiMP-50}$-RGBT} & 0.767 & 0.534 & 0.639 & 0.529 \\  
			\hline              
		\end{tabular}%
	}
	\centering
	\label{tb::ablation1}
\end{table}

\subsection{Evaluation on Synthetic Data}
To further validate the effectiveness of our MArMOT model, we construct a synthetic RGB-thermal cross-modal dataset from existing RGBT datasets, including GTOT~\cite{li2016learning} and RGBT234 ~\cite{li2019rgb}. 
In order to simulate cross-modal tracking tasks more accurately, each frame in synthetic videos is generated by selecting one modality from corresponding RGB and thermal images, according to the challenge labels of illumination various and thermal crossover.
Specifically, if there is no illumination various in the first frame, each sequence always starts with the RGB modality, and the modality switch is only performed when the challenge of illumination various or thermal crossover occurs. In addition, if there are too many modality switch (i.e., the number of modality-switch is greater than 5), the sequence is considered to be too challenging and the sequence is discarded; if there is no modality switch in the entire sequence, then part of 1/4 to 1/2 in current sequence will switch to another modality.

For this synthetic dataset, we select the RGBT234 dataset as the training set and GTOT dataset as the testing set, and retrain the entire network for the cross-modal RGBT tracking task with proposed three-stage training method. The experimental results are shown in Table~\ref{tb::ablation1}.

From the results we can see that our MArMOT model can well handle appearance gap between RGB and thermal modalities in tracking process, and thus further prove the generalization and effectiveness of our method in handling the different cross-modal tracking task.

\begin{table}[htp]
	\centering
	\caption{Comparison of several variants of our MArMOT, where * indicates the tracker is re-trained using CMOTB training dataset. }
	\resizebox{.5\textwidth}{!}{%
		\begin{tabular}{c|cccc}
			\hline
			Trackers & PR & NPR & SR-I & SR-II  \\
			\hline
			\textbf{DiMP-50} & 0.573 & 0.566  & 0.628 & 0.536\\
			\textbf{DiMP-50*} & 0.642 & 0.630 & 0.710 & 0.599\\
			\textbf{MArMOT$\bf_{DiMP-50}$-one-stage} & 0.662 & 0.646 &0.726 & 0.612\\
			\hline   
			\textbf{MArMOT$\bf_{DiMP-50}$} & 0.685 & 0.663 & 0.746 & 0.628 \\  
			\hline              
		\end{tabular}%
	}
	\centering
	\label{tb::ablation2}
\end{table}

\subsection{Effectiveness of Modality-aware Representations}
To verify the effectiveness of the proposed modality-aware representations and proposed three-stage training method, we implement the variant trackers, named MArMOT$\bf_{DiMP-50}$-one-stage, by using one-stage training method to train the entire networks together on the CMOTB dataset. 

The results are shown in Table~\ref{tb::ablation2}. The experimental results show that proposed three-stage training method outperforms the one-stage training method with $2.3\%$/$1.7\%$/$2.0\%$/$1.6\%$ gains in PR/NPR/SR-I/SR-II, which can prove that the proposed three-stage learning method is benefit to modality-aware branches to learn corresponding modality-specific target representations. 
In addition, we can also find that the performance of the one-stage training method is still better than DiMP-50*, which can verify the effectiveness of the proposed MArMOT model for mining cross-modal information.
% \begin{table}[htp]
% 	\centering
% 	\caption{Comparison of several variants of our MArMOT, where * indicates the tracker is re-trained using CMOTB training dataset. }
% 	\resizebox{.5\textwidth}{!}{%
% 		\begin{tabular}{c|cccc}
% 			\hline
% 			Trackers & NPR & PR & SR-I & SR-II  \\
% 			\hline
% 			\textbf{RT-MDNet} & 0.426 & 0.433  & 0.425 & 0.354\\
% 			\textbf{RT-MDNet+MU} & 0.459 & 0.457 & 0.448 & 0.379\\
% 			\textbf{RT-MDNet+MU}* & 0.470 & 0.484 &0.473 & 0.397\\
% 			MArMOT-MU & 0.571 & 0.559 & 0.591 & 0.487\\
% 			\hline   
% 			\textbf{MArMOT}* & 0.592 & 0.576 & 0.639 & 0.507 \\  
% 			\hline              
% 		\end{tabular}%
% 	}
% 	\centering
% 	\label{tb::ablation}
% \end{table}

% \subsection{Ablation Study}
% % To verify the effectiveness of main components of the proposed tracker, we carry out the ablation study on the CMOTB testing dataset. We compare three variants against the baseline RT-MDNet\cite{jung2018real}, and they are: 1) RT-MDNet+MU, that adds the meta updater in online update in RT-MDNet. 2) RT-MDNet+MU*, that is re-trained using CMOTB training dataset. 3) MArMOT-MU, that removes the meta updater in online update in MArMOT. The results are shown in Table~\ref{tb::ablation}, and we find that the meta updater indeed improves tracking performance and our network and learning algorithm play a core role in performance boosting by observing that MArMOT-MU is significantly superior than RT-MDNet\cite{jung2018real}. 
% pass

\section{Conclusion}
We provide a large-scale cross-modal object tracking benchmark with high-quality dense bounding box annotations. And we also propose a simple yet effective method based on a modality-aware feature learning algorithm for cross-modal object tracking purpose. Extensive experiments on the dataset demonstrate the effectiveness of the proposed method against state-of-the-art trackers. By releasing this dataset, we believe that it will help the research and developments of cross-modal object tracking. In the future, we will study more effective tracking algorithms to solve the cross-modal tracking problem and extend the dataset to cover more real-world scenarios.
\bibliography{aaai22}

\begin{thebibliography}{27}
\providecommand{\natexlab}[1]{#1}

\bibitem[{Bertinetto et~al.(2016)Bertinetto, Valmadre, Henriques, Vedaldi, and
  Torr}]{bertinetto2016fully}
Bertinetto, L.; Valmadre, J.; Henriques, J.~F.; Vedaldi, A.; and Torr, P.~H.
  2016.
\newblock Fully-convolutional siamese networks for object tracking.
\newblock In \emph{Proceedings of the European Conference on Computer Vision},
  850--865. Springer.

\bibitem[{Bhat et~al.(2019)Bhat, Danelljan, Gool, and
  Timofte}]{bhat2019learning}
Bhat, G.; Danelljan, M.; Gool, L.~V.; and Timofte, R. 2019.
\newblock Learning discriminative model prediction for tracking.
\newblock In \emph{Proceedings of the IEEE International Conference on Computer
  Vision}, 6182--6191.

\bibitem[{Biresaw et~al.(2016)Biresaw, Nawaz, Ferryman, and
  Dell}]{biresaw2016vitbat}
Biresaw, T.~A.; Nawaz, T.; Ferryman, J.; and Dell, A.~I. 2016.
\newblock Vitbat: Video tracking and behavior annotation tool.
\newblock In \emph{Proceedings of the IEEE International Conference on Advanced
  Video and Signal Based Surveillance}, 295--301. IEEE.

\bibitem[{Chen et~al.(2020)Chen, Zhong, Li, Zhang, and Ji}]{chen2020siamese}
Chen, Z.; Zhong, B.; Li, G.; Zhang, S.; and Ji, R. 2020.
\newblock Siamese box adaptive network for visual tracking.
\newblock In \emph{Proceedings of the IEEE Conference on Computer Vision and
  Pattern Recognition}, 6668--6677.

\bibitem[{Choi, Kwon, and Lee(2020)}]{choi2020visual}
Choi, J.; Kwon, J.; and Lee, K.~M. 2020.
\newblock Visual tracking by tridentalign and context embedding.
\newblock In \emph{Proceedings of the Asian Conference on Computer Vision}.

\bibitem[{Dai et~al.(2020)Dai, Zhang, Wang, Li, Lu, and Yang}]{dai2020high}
Dai, K.; Zhang, Y.; Wang, D.; Li, J.; Lu, H.; and Yang, X. 2020.
\newblock High-performance long-term tracking with meta-updater.
\newblock In \emph{Proceedings of the IEEE Conference on Computer Vision and
  Pattern Recognition}, 6298--6307.

\bibitem[{Danelljan et~al.(2019)Danelljan, Bhat, Khan, and
  Felsberg}]{danelljan2019atom}
Danelljan, M.; Bhat, G.; Khan, F.~S.; and Felsberg, M. 2019.
\newblock Atom: Accurate tracking by overlap maximization.
\newblock In \emph{Proceedings of the IEEE Conference on Computer Vision and
  Pattern Recognition}, 4660--4669.

\bibitem[{Huang, Zhao, and Huang(2020)}]{huang2020globaltrack}
Huang, L.; Zhao, X.; and Huang, K. 2020.
\newblock Globaltrack: A simple and strong baseline for long-term tracking.
\newblock In \emph{Proceedings of the AAAI Conference on Artificial
  Intelligence}, volume~34, 11037--11044.

\bibitem[{Jung et~al.(2018)Jung, Son, Baek, and Han}]{jung2018real}
Jung, I.; Son, J.; Baek, M.; and Han, B. 2018.
\newblock Real-time mdnet.
\newblock In \emph{Proceedings of the European Conference on Computer Vision},
  83--98.

\bibitem[{Kohavi et~al.(1995)}]{kohavi1995study}
Kohavi, R.; et~al. 1995.
\newblock A study of cross-validation and bootstrap for accuracy estimation and
  model selection.
\newblock In \emph{Proceedings of the International Joint Conference on
  Artificial Intelligence}, volume~14, 1137--1145. Montreal, Canada.

\bibitem[{Li et~al.(2019{\natexlab{a}})Li, Wu, Wang, Zhang, Xing, and
  Yan}]{li2019siamrpn++}
Li, B.; Wu, W.; Wang, Q.; Zhang, F.; Xing, J.; and Yan, J. 2019{\natexlab{a}}.
\newblock Siamrpn++: Evolution of siamese visual tracking with very deep
  networks.
\newblock In \emph{Proceedings of the IEEE Conference on Computer Vision and
  Pattern Recognition}, 4282--4291.

\bibitem[{Li et~al.(2016)Li, Cheng, Hu, Liu, Tang, and Lin}]{li2016learning}
Li, C.; Cheng, H.; Hu, S.; Liu, X.; Tang, J.; and Lin, L. 2016.
\newblock Learning collaborative sparse representation for grayscale-thermal
  tracking.
\newblock \emph{IEEE Transactions on Image Processing}, 25(12): 5743--5756.

\bibitem[{Li et~al.(2019{\natexlab{b}})Li, Liang, Lu, Zhao, and
  Tang}]{li2019rgb}
Li, C.; Liang, X.; Lu, Y.; Zhao, N.; and Tang, J. 2019{\natexlab{b}}.
\newblock RGB-T object tracking: benchmark and baseline.
\newblock \emph{Pattern Recognition}, 96: 106977.

\bibitem[{Li et~al.(2020)Li, Liu, Lu, Ji, and Tang}]{li2020challenge}
Li, C.; Liu, L.; Lu, A.; Ji, Q.; and Tang, J. 2020.
\newblock Challenge-aware rgbt tracking.
\newblock In \emph{Proceedings of the European Conference on Computer Vision},
  222--237. Springer.

\bibitem[{Li et~al.(2019{\natexlab{c}})Li, Chen, Ouyang, Wang, Yang, and
  Lu}]{li2019gradnet}
Li, P.; Chen, B.; Ouyang, W.; Wang, D.; Yang, X.; and Lu, H.
  2019{\natexlab{c}}.
\newblock GradNet: Gradient-guided network for visual object tracking.
\newblock In \emph{Proceedings of the IEEE International Conference on Computer
  Vision}, 6162--6171.

\bibitem[{Li et~al.(2019{\natexlab{d}})Li, Wang, Hu, and
  Yang}]{li2019selective}
Li, X.; Wang, W.; Hu, X.; and Yang, J. 2019{\natexlab{d}}.
\newblock Selective kernel networks.
\newblock In \emph{Proceedings of the IEEE Conference on Computer Vision and
  Pattern Recognition}, 510--519.

\bibitem[{Muller et~al.(2018)Muller, Bibi, Giancola, Alsubaihi, and
  Ghanem}]{muller2018trackingnet}
Muller, M.; Bibi, A.; Giancola, S.; Alsubaihi, S.; and Ghanem, B. 2018.
\newblock Trackingnet: A large-scale dataset and benchmark for object tracking
  in the wild.
\newblock In \emph{Proceedings of the European Conference on Computer Vision},
  300--317.

\bibitem[{Nam and Han(2016)}]{nam2016learning}
Nam, H.; and Han, B. 2016.
\newblock Learning multi-domain convolutional neural networks for visual
  tracking.
\newblock In \emph{Proceedings of the IEEE Conference on Computer Vision and
  Pattern Recognition}, 4293--4302.

\bibitem[{Song and Xiao(2013)}]{Song13iccv}
Song, S.; and Xiao, J. 2013.
\newblock Tracking revisited using RGBD camera: Unified benchmark and
  baselines.
\newblock In \emph{Proceedings of the IEEE International Conference on Computer
  Vision}, 233--240.

\bibitem[{Song et~al.(2018)Song, Ma, Wu, Gong, Bao, Zuo, Shen, Lau, and
  Yang}]{song2018vital}
Song, Y.; Ma, C.; Wu, X.; Gong, L.; Bao, L.; Zuo, W.; Shen, C.; Lau, R.~W.; and
  Yang, M.-H. 2018.
\newblock Vital: Visual tracking via adversarial learning.
\newblock In \emph{Proceedings of the IEEE Conference on Computer Vision and
  Pattern Recognition}, 8990--8999.

\bibitem[{Szegedy et~al.(2016)Szegedy, Ioffe, Vanhoucke, and
  Alemi}]{szegedy2016inception}
Szegedy, C.; Ioffe, S.; Vanhoucke, V.; and Alemi, A. 2016.
\newblock Inception-v4, inception-resnet and the impact of residual connections
  on learning.
\newblock \emph{arXiv preprint arXiv:1602.07261}.

\bibitem[{Van~der Maaten and Hinton(2008)}]{van2008visualizing}
Van~der Maaten, L.; and Hinton, G. 2008.
\newblock Visualizing data using t-SNE.
\newblock \emph{Journal of Machine Learning Research}, 9(11).

\bibitem[{Wang et~al.(2019)Wang, Zhang, Bertinetto, Hu, and
  Torr}]{wang2019fast}
Wang, Q.; Zhang, L.; Bertinetto, L.; Hu, W.; and Torr, P.~H. 2019.
\newblock Fast online object tracking and segmentation: A unifying approach.
\newblock In \emph{Proceedings of the IEEE Conference on Computer Vision and
  Pattern Recognition}, 1328--1338.

\bibitem[{Yan et~al.(2019)Yan, Zhao, Wang, Lu, and Yang}]{yan2019skimming}
Yan, B.; Zhao, H.; Wang, D.; Lu, H.; and Yang, X. 2019.
\newblock 'Skimming-Perusal'Tracking: A framework for real-time and robust
  long-term tracking.
\newblock In \emph{Proceedings of the IEEE International Conference on Computer
  Vision}, 2385--2393.

\bibitem[{Zhang and Peng(2019)}]{zhang2019deeper}
Zhang, Z.; and Peng, H. 2019.
\newblock Deeper and wider siamese networks for real-time visual tracking.
\newblock In \emph{Proceedings of the IEEE Conference on Computer Vision and
  Pattern Recognition}, 4591--4600.

\bibitem[{Zhang et~al.(2020)Zhang, Peng, Fu, Li, and Hu}]{zhang2020ocean}
Zhang, Z.; Peng, H.; Fu, J.; Li, B.; and Hu, W. 2020.
\newblock Ocean: Object-aware anchor-free tracking.
\newblock In \emph{Proceedings of the European Conference on Computer Vision},
  771--787. Springer.

\bibitem[{Zhu et~al.(2018)Zhu, Wang, Li, Wu, Yan, and Hu}]{zhu2018distractor}
Zhu, Z.; Wang, Q.; Li, B.; Wu, W.; Yan, J.; and Hu, W. 2018.
\newblock Distractor-aware siamese networks for visual object tracking.
\newblock In \emph{Proceedings of the European Conference on Computer Vision},
  101--117.

\end{thebibliography}
\end{document}